\documentclass{article}
\usepackage{spconf,amsmath,graphicx,hyperref}
\usepackage{amsmath,amssymb,amsfonts}
\usepackage{subcaption}
\usepackage{graphicx}

\usepackage{multirow} 
\usepackage{caption}
\usepackage{subcaption}
\usepackage{pifont}
\usepackage{amsmath}
\usepackage{amssymb} 
\usepackage{booktabs}

\title{Shared DIFF Transformer}
\name{Yueyang Cang$^{\star,1}$\thanks{1 Equal contribution} \qquad Yuhang Liu$^{\star,1}$ \qquad Xiaoteng Zhang$^{\star}$ \qquad Li Shi$^{\star}$ \qquad Wenge Que$^{\ddagger}$}

\address{
  $^{\star}$ Tsinghua University, Beijing, China \\
  $^{\ddagger}$ Donghua University, Shanghai, China
}

\begin{document}
%
\maketitle
\begin{abstract}
DIFF Transformer improves attention allocation by enhancing relevant context focus while suppressing noise through a differential mechanism computing differences between two independent attention distributions, though its independent signal generation causes parameter redundancy and unstable noise extraction. To address this, we propose Shared DIFF Transformer that introduces a shared base matrix for global pattern modeling combined with low-rank updates for task-specific flexibility, significantly reducing redundancy while improving efficiency and noise extraction stability. Experiments demonstrate our method's superior and more stable performance in long-sequence modeling, key information retrieval, and in-context learning compared to DIFF Transformer, offering a novel solution for optimizing differential attention mechanisms in robust Transformer architectures.
\end{abstract}
\begin{keywords}
Shared Differential Attention, Transformer, Long-context Modeling, In-context Learning
\end{keywords}
\section{Introduction}
\label{sec:intro}

Transformers have achieved significant success in tasks like natural language processing and vision, thanks to their powerful self-attention mechanism. However, standard Transformers often overallocate attention to irrelevant context, leading to inefficiencies, especially in long-context modeling and key information retrieval. To address this, DIFF Transformer introduces a differential attention mechanism inspired by noise-canceling headphones, which amplifies relevant context and cancels noise, optimizing attention and forming sparse patterns. While this improves focus on critical inputs, it also introduces parameter redundancy and unstable noise extraction, leaving room for further optimization in model complexity, efficiency, and stability.

In this study, we propose Shared DIFF Transformer, inspired by a differential amplifier, which uses a shared base matrix to model global patterns and low-rank updates for task-specific flexibility. This reduces parameter redundancy, improves efficiency, and enhances robustness in complex scenarios. Extensive experiments across various tasks show that Shared DIFF Transformer achieves comparable language modeling performance to DIFF Transformer, while significantly reducing parameters and training tokens. It also outperforms in downstream tasks, demonstrating excellent scalability and positioning Shared DIFF Transformer as a robust and efficient architecture for large-scale language models.

\section{Shared Differential Transformer}
\label{sec:format}


\begin{figure*}[h]
    \centering
    \begin{minipage}[b]{0.48\textwidth}
        \centering
        \includegraphics[width=0.7\textwidth]{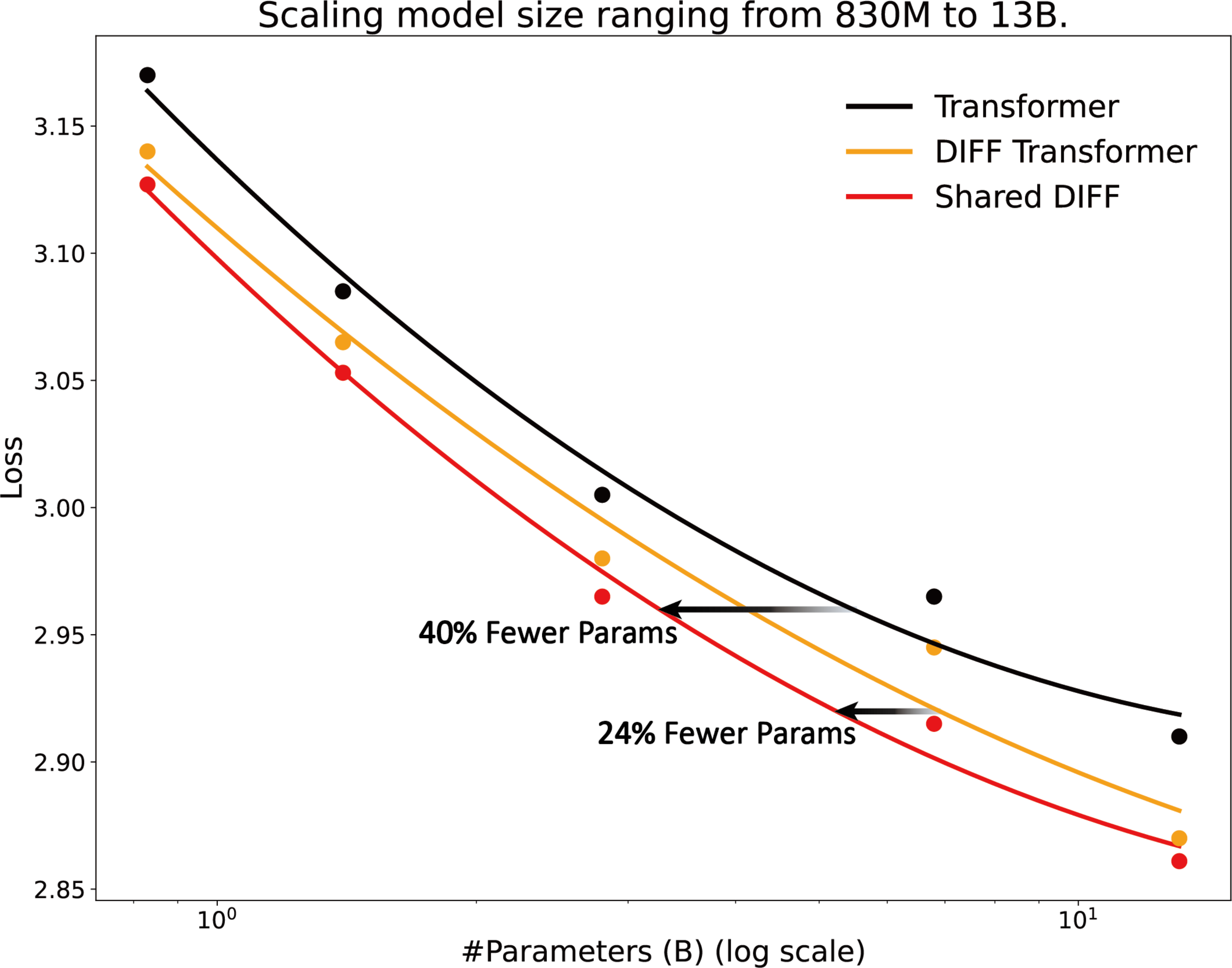}
        \caption*{(a) Scaling model size from 830M to 13B.}
    \end{minipage}
    \hfill
    \begin{minipage}[b]{0.48\textwidth}
        \centering
        \includegraphics[width=0.7\textwidth]{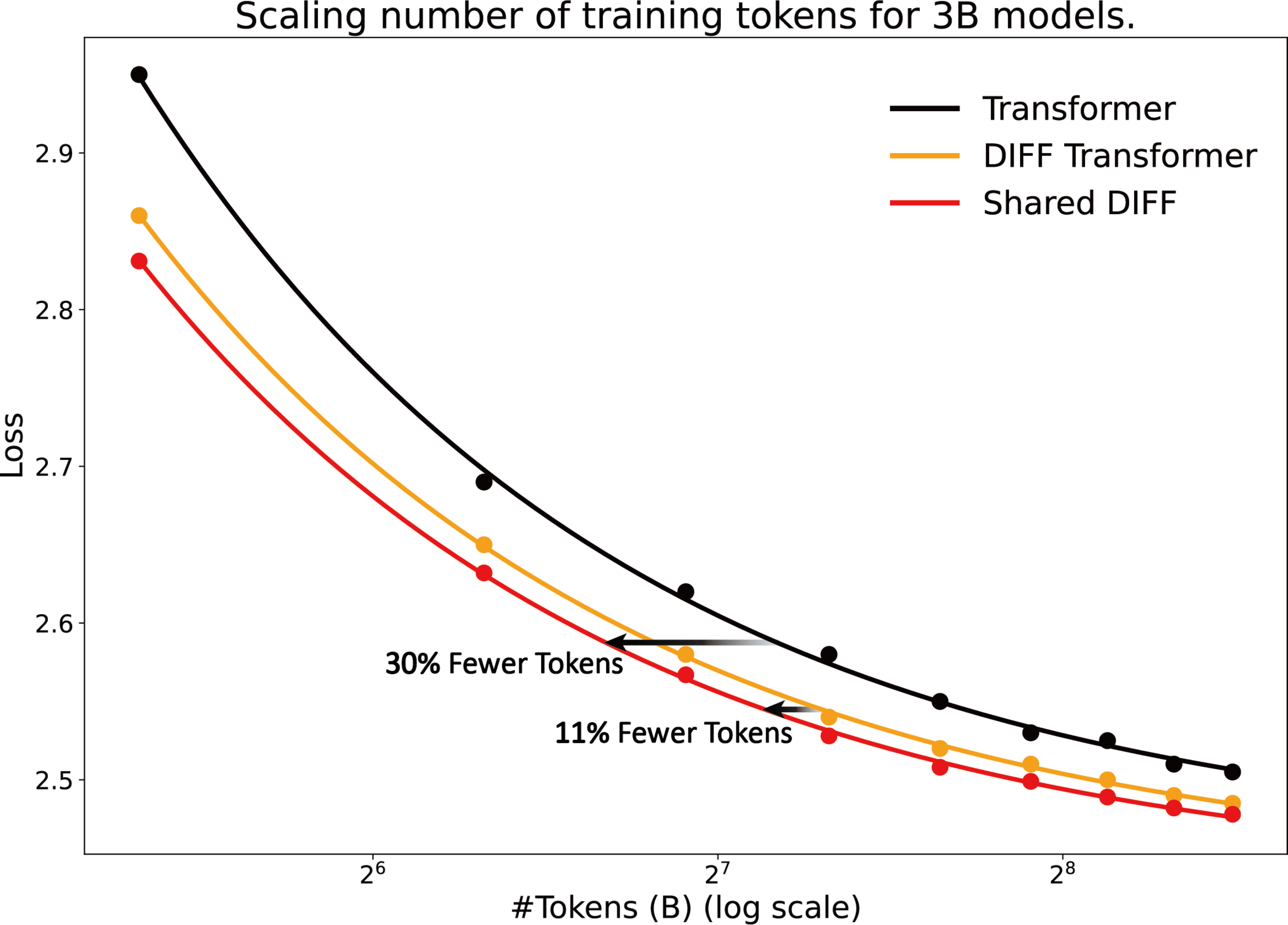}
        \caption*{(b) Scaling the number of training tokens for the 3B model.}
    \end{minipage}
    \caption{Language modeling loss with scaled model size and training tokens. (a) Shared DIFF Transformer achieves comparable performance to larger models with fewer parameters. (b) It achieves similar performance with fewer training tokens.}
    \label{fig:scaling_comparison}
\end{figure*}

In Shared Differential Transformer, the shared differential attention mechanism is used to improve the original attention mechanism.The shared differential attention mechanism maps query, key, and value vectors to outputs, leveraging a shared base matrix to model global patterns and low-rank updates for task-specific refinements. Specifically, given an input $X \in \mathbb{R}^{N \times d_\text{model}}$, we first project it to query, key, and value matrices as follows:
\begin{equation}
Q_1 = X W_{Q1}, \quad Q_2 = X W_{Q2},
\end{equation}
\begin{equation}
K_1 = X W_{K1}, \quad K_2 = X W_{K2}, \quad V = X W_V,
\end{equation}

where $W_{Q1}, W_{Q2}, W_{K1}, W_{K2} \in \mathbb{R}^{d_\text{model} \times d}$ and $W_V \in \mathbb{R}^{d_\text{model} \times 2d}$.

To reduce parameter redundancy and enhance model flexibility, $W_{Q1}, W_{Q2}, W_{K1}, W_{K2}$ are redefined using a shared base matrix and low-rank updates:
\begin{equation}
\small
{\scriptsize
W_{Q1} = W_Q + W_{q11} W_{q12}^\top, \quad W_{Q2} = W_Q + W_{q21} W_{q22}^\top
}
\end{equation}
\begin{equation}
\small
{\scriptsize
W_{K1} = W_K + W_{k11} W_{k12}^\top, \quad W_{K2} = W_K + W_{k21} W_{k22}^\top
}
\end{equation}

where $W_Q$ and $W_K \in \mathbb{R}^{d_\text{model} \times d}$ are shared base matrices used to capture global patterns. $W_{qi1}$ and $W_{ki1} \in \mathbb{R}^{d_\text{model} \times r}$, as well as $W_{qi2}$ and $W_{ki2} \in \mathbb{R}^{d \times r}$, are introduced as low-rank matrices for dynamic adjustments. These low-rank updates allow the model to adapt flexibly to different contexts while preserving shared global information, ensuring parameter efficiency and enhancing task-specific expressiveness.

The attention scores are computed as:
\begin{equation}
\small
A_1 = \text{softmax}\left(\frac{Q_1 K_1^\top}{\sqrt{d}}\right), \quad A_2 = \text{softmax}\left(\frac{Q_2 K_2^\top}{\sqrt{d}}\right).
\end{equation}
The final shared differential attention is defined as:
\[
\text{SharedDiffAttn}(X) = (A_1 - \lambda A_2) V,
\]
where $\lambda$ is a learnable scalar controlling the contribution of $A_2$. To stabilize learning dynamics, $\lambda$ is re-parameterized as:
\[
\lambda = \exp(\lambda_{q1} \cdot \lambda_{k1}) - \exp(\lambda_{q2} \cdot \lambda_{k2}) + \lambda_\text{init},
\]
where $\lambda_{q1}, \lambda_{k1}, \lambda_{q2}, \lambda_{k2} \in \mathbb{R}^d$ are learnable vectors, and $\lambda_\text{init} \in (0, 1)$ is a constant for initialization. This reparameterization ensures consistent training dynamics, allowing us to effectively inherit hyperparameters from standard Transformers without extensive tuning.

The introduction of shared base matrices significantly reduces parameter complexity compared to DIFF Transformer. Specifically, the parameter count for query and key projections is reduced from $4 \cdot d_\text{model} \cdot d$ to $2 \cdot d_\text{model} \cdot d + 2 \cdot d_\text{model} \cdot r + 2 \cdot d \cdot r$, where $r$ is the rank of the low-rank updates ($r \ll d, d_\text{model}$).

\section{Experiments}
\label{sec:pagestyle}

We evaluate Shared DIFF Transformer by comparing it with DIFF Transformer across various downstream tasks, model scaling, and training tokens (Sections 3.1 and 3.2). We also highlight its advantages in key information retrieval and in-context learning (Sections 3.3 and 3.4) and conduct ablation studies to assess the impact of design choices on performance (Section 3.5).

\subsection{Language Modeling Evaluation}

\textbf{Setup.} We trained a 3B-size Shared DIFF Transformer language model and compared it with the 3B-size DIFF Transformer.

\textbf{Results.} Table \ref{tab:results} presents zero-shot evaluation results on the LM Eval Harness benchmark \cite{gao2021framework}, showing mean scores and standard deviations across five runs. We compare Shared DIFF Transformer with state-of-the-art models such as OpenLLaMA-v2-3B \cite{geng2023openllama}, StableLM-base-alpha-3B-v2 \cite{tow2023stablelm}, and StableLM-3B-4E1T \cite{stableLM}, all trained on 1 trillion tokens under identical conditions. Shared DIFF Transformer outperforms these models across all tasks, demonstrating superior ability to capture both local and global dependencies.

Notably, Shared DIFF Transformer shows consistent improvements, particularly with the integral mechanism, which enhances the use of global information and leads to exceptional performance on challenging benchmarks like ARC-C, BoolQ, and PIQA.

\begin{table*}[h]
    \centering
    \resizebox{\textwidth}{!}{%
    \begin{tabular}{|l|c|c|c|c|c|c|c|c|}
    \hline
    \textbf{Model} & \textbf{ARC-C} & \textbf{ARC-E} & \textbf{BoolQ} & \textbf{HellaSwag} & \textbf{OBQA} & \textbf{PIQA} & \textbf{WinoGrande} & \textbf{Avg} \\ \hline
    OpenLLaMA-3B-v2  & 33.9 & 67.6 & 65.7 & 70.0 & 26.6 & 76.7 & 62.9 & 57.5 \\ 
    StableLM-base-alpha-3B-v2  & 32.4 & 67.3 & 64.6 & 68.6 & 27.1 & 76.0 & 63.0 & 57.0 \\ 
    StableLM-3B-4E1T & -- & 66.6 & -- & -- & 25.5 & 76.8 & 63.2 & -- \\ 
    DIFF-3B & 36.9\,\textpm\,2.1 & 72.6\,\textpm\,1.7 & 69.2\,\textpm\,1.8 & 71.1\,\textpm\,2.4 & 29.1\,\textpm\,0.8 & 76.5\,\textpm\,1.0 & 69.2\,\textpm\,2.0 & 60.6 \\ 
    \textbf{Shared DIFF-3B} & \textbf{39.8\,\textpm\,1.2} & \textbf{74.2\,\textpm\,0.9} & \textbf{71.2\,\textpm\,1.4} & \textbf{72.5\,\textpm\,1.7} & \textbf{30.8\,\textpm\,0.4} & \textbf{78.8\,\textpm\,0.9} & \textbf{70.5\,\textpm\,1.2} & \textbf{62.5} \\ \hline
    \end{tabular}%
    }
    \caption{Eval Harness  accuracy compared with well-trained Transformer language models. The results indicate the superior performance of Shared DIFF Transformer over other models across a range of tasks.}
    \label{tab:results}
\end{table*}

\subsection{Scalability Compared with Transformer}

We evaluated the scalability of Shared DIFF Transformer against the standard Transformer, focusing on language modeling tasks.

\textbf{Scaling Model Size} As shown in Figure \ref{fig:scaling_comparison}(a), Shared DIFF Transformer outperformed both Transformer and DIFF Transformer across various model sizes, achieving similar validation loss to Transformer with 40\% fewer parameters, and matching DIFF Transformer with 24\% fewer parameters. This demonstrates its superior parameter efficiency and scalability.

\textbf{Scaling Training Tokens} Figure \ref{fig:scaling_comparison}(b) shows that Shared DIFF Transformer achieved comparable performance to Transformer with 30\% fewer training tokens, and surpassed DIFF Transformer with 11\% fewer tokens. These results highlight its significant data efficiency and ability to perform equally or better with fewer resources.

\subsection{Key Information Retrieval}

The Needle-In-A-Haystack test \cite{kamradt2023needle} measures how well models identify key information in large contexts. "Needles" are short sentences linking a city to a unique identifier, and the goal is to retrieve them accurately. The correct needle is placed at different positions (0\%, 25\%, 50\%, 75\%, 100\%), with others randomly placed. Each context and needle position combination is evaluated over 50 samples, and the average accuracy is reported.

\begin{table}[h]
    \centering
    \resizebox{0.45\textwidth}{!}{%
    \begin{tabular}{|l|c|c|c|c|}
        \hline
        \multirow{2}{*}{\textbf{Model}} & \textbf{$N = 1$} & \textbf{$N = 2$} & \textbf{$N = 4$} & \textbf{$N = 6$} \\ 
                                      & \textbf{$R = 1$} & \textbf{$R = 2$} & \textbf{$R = 2$} & \textbf{$R = 2$} \\ \hline
        Transformer                   & \textbf{1.00}    & 0.85            & 0.62            & 0.55            \\ 
        DIFF                           & \textbf{1.00}    & 0.92            & 0.84            & 0.85            \\ 
        Shared DIFF                    & \textbf{1.00}    & \textbf{0.95}    & \textbf{0.89}    & \textbf{0.87}    \\ \hline
    \end{tabular}%
    }
    \caption{Multi-needle retrieval accuracy in 4K-length contexts, averaged over the answer needle positions. $N$ represents the number of needles, and $R$ denotes the number of query cities.}
    \label{tab:multi_needle_retrieval}
\end{table}

\textbf{Results from 4K Contexts} Using 4K-length contexts and varying needle counts ($N = 1, 2, 4, 6$) and retrieval counts ($R = 1, 2$), Shared DIFF Transformer outperformed both Transformer and DIFF models, especially as needle and query city counts increased. For $N = 6$ and $R = 2$, Shared DIFF Transformer achieved 0.87 accuracy, surpassing other models, demonstrating its efficiency in extracting relevant information amid irrelevant data.

\textbf{Retrieve from 64K Context Length} With extended 64K context lengths ($N = 8$, $R = 1$), Shared DIFF Transformer outperformed Transformer and DIFF Transformer. At the 25\% depth in a 40K context, it showed 48\% improvement over Transformer and 8\% over DIFF Transformer in accuracy.

\begin{figure}[h!]
    \centering
    \includegraphics[width=\linewidth]{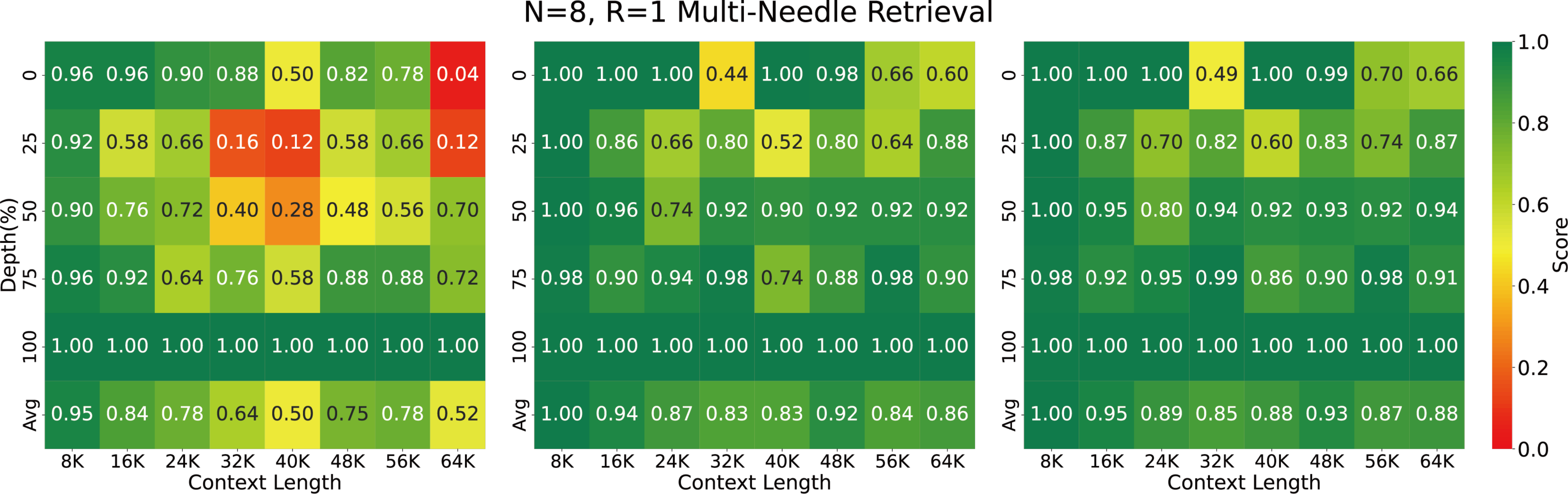}  
    \caption{Multi-needle retrieval results in 64K length.}
    \label{fig:multi_needle_results}
\end{figure}

\textbf{Attention Score Analysis} Table \ref{tab:attention_dint} compares the attention scores for the correct answer and irrelevant context in the retrieval task. Shared DIFF Transformer assigns higher attention to the correct answer and less to irrelevant context, especially in early depths (0\%, 25\%, and 50\%).

\begin{table*}[h]
\centering
\begin{tabular}{lcccccccccc}
\toprule
\textbf{Model} & \multicolumn{5}{c}{\textbf{Attention to Answer↑}} & \multicolumn{5}{c}{\textbf{Attention Noise↓}} \\
 & 0\% & 25\% & 50\% & 75\% & 100\% & 0\% & 25\% & 50\% & 75\% & 100\% \\
\midrule
Transformer & 0.03 & 0.03 & 0.03 & 0.07 & 0.09 & 0.51 & 0.54 & 0.52 & 0.49 & 0.49 \\
DIFF & 0.27 & 0.30 & 0.31 & 0.32 & 0.40 & 0.01 & 0.02 & 0.02 & 0.02 & 0.01 \\
Shared DIFF & \textbf{0.33} & \textbf{0.36} & \textbf{0.39} & \textbf{0.41} & \textbf{0.44} & \textbf{0.01} & \textbf{0.01} & \textbf{0.02} & \textbf{0.01} & \textbf{0.01} \\
\bottomrule
\end{tabular}
\caption{Attention scores in the key information retrieval task, with the target answer placed at varying depths. Shared DIFF Transformer allocates more attention to relevant information and reduces attention to noise.}
\label{tab:attention_dint}
\end{table*}

\subsection{In-Context Learning}

We examine in-context learning from two aspects: its effectiveness in many-shot classification and the model's ability to maintain robustness with context. In-context learning is a crucial feature of language models, showcasing their efficiency in using input context.

\begin{figure}[h!]
    \centering
    \begin{minipage}{0.23\textwidth}
        \centering
        \includegraphics[width=\linewidth]{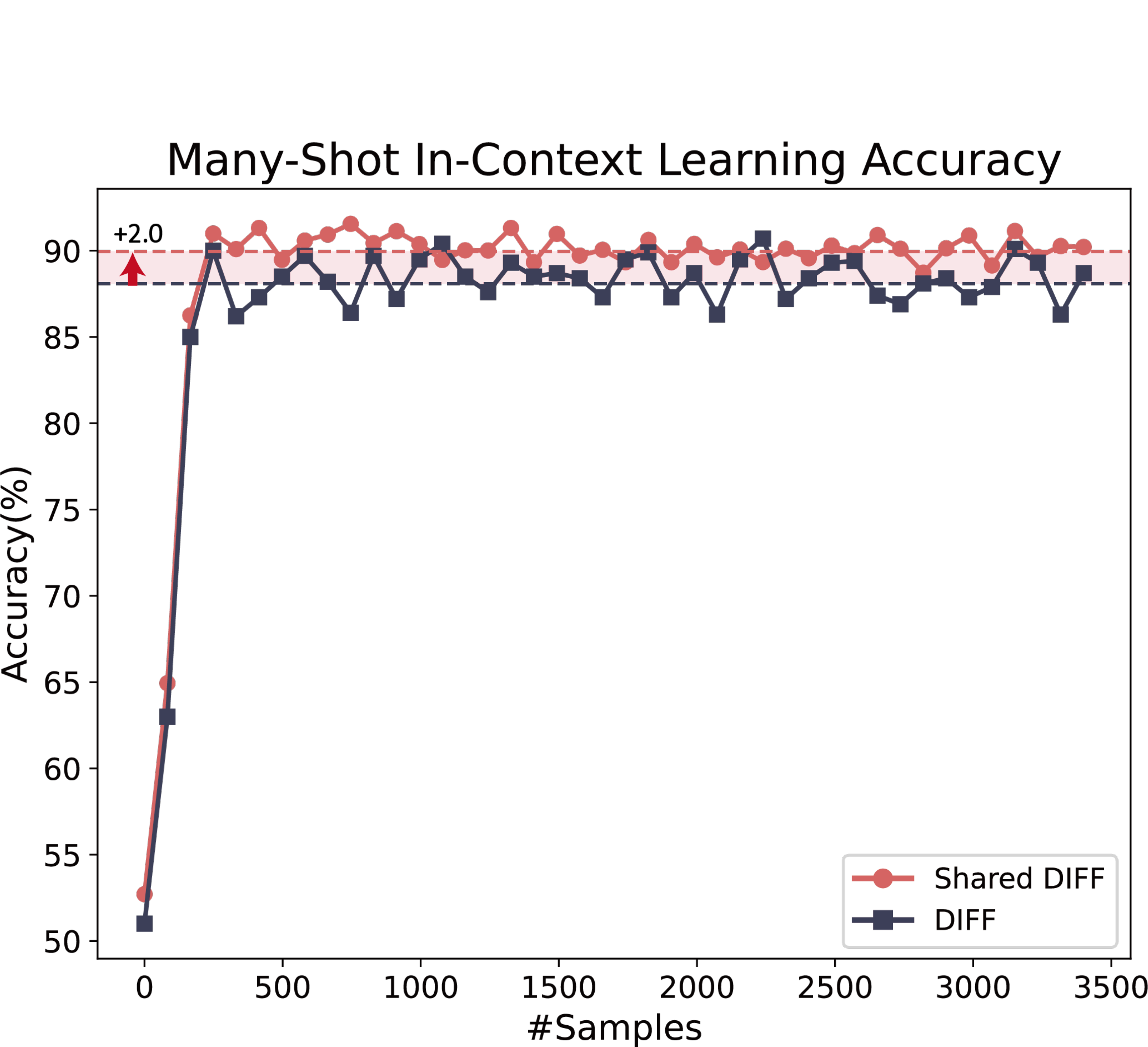} 
        \subcaption{TREC dataset} \label{fig:trec}
    \end{minipage}%
    \hspace{0.3cm} 
    \begin{minipage}{0.23\textwidth}
        \centering
        \includegraphics[width=\linewidth]{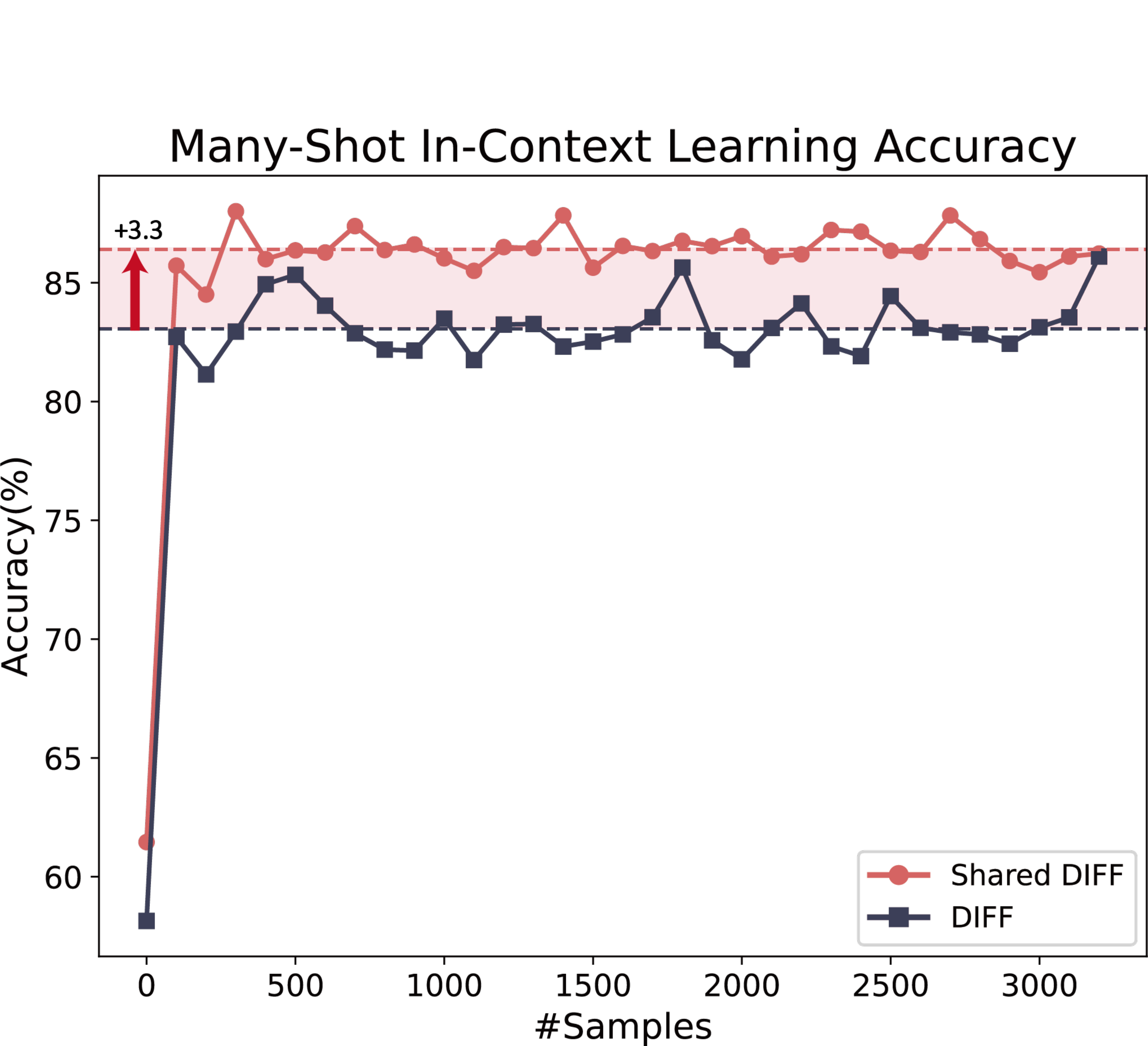} 
        \subcaption{TREC-fine dataset} \label{fig:trec_fine}
    \end{minipage}%
    \hspace{0.3cm} 
    \begin{minipage}{0.23\textwidth}
        \centering
        \includegraphics[width=\linewidth]{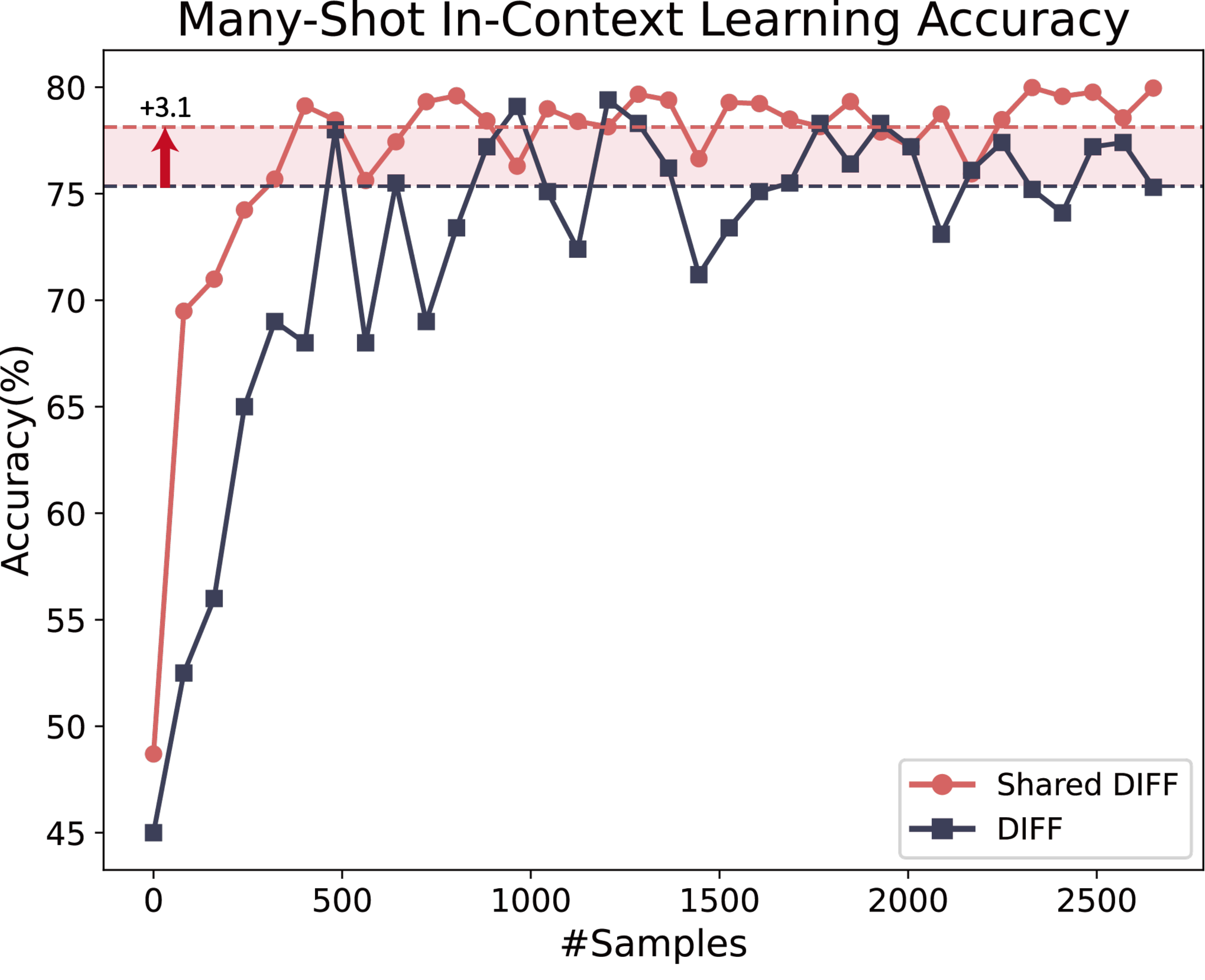} 
        \subcaption{Banking-77 dataset} \label{fig:banking_77}
    \end{minipage}%
    \hspace{0.3cm} 
    \begin{minipage}{0.23\textwidth}
        \centering
        \includegraphics[width=\linewidth]{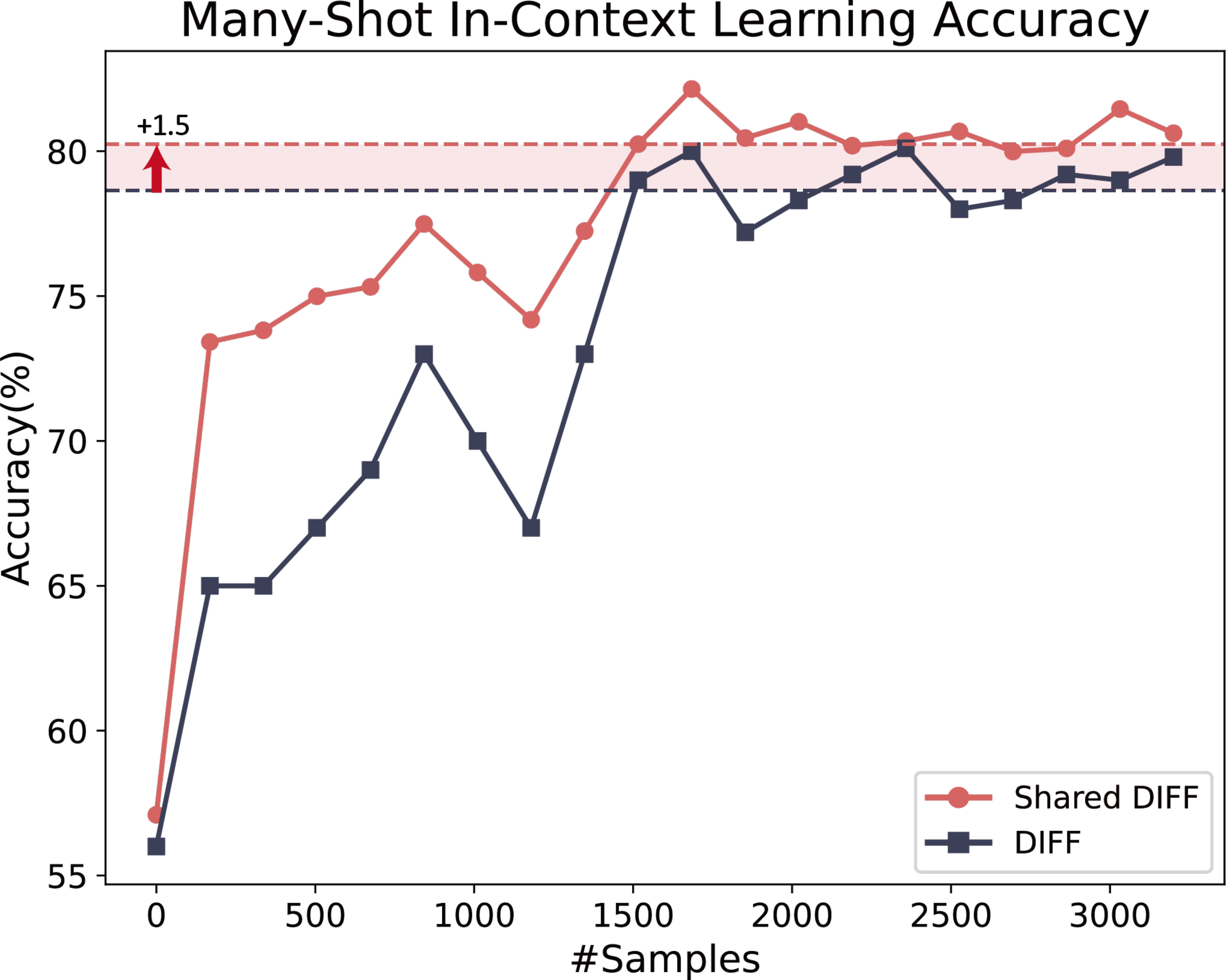} 
        \subcaption{Clinic-150 dataset} \label{fig:clinic_150}
    \end{minipage}
    \caption{Accuracy of many-shot in-context learning across four datasets, with demonstration examples increasing from 1-shot up to a total of 64K tokens. The dashed lines indicate the average accuracy once the model's performance stabilizes.} 
    \label{fig:many_shot_incontext_learning}
\end{figure}

\begin{figure}[h!]
    \centering
    \begin{minipage}[b]{0.48\linewidth}
        \centering
        \includegraphics[width=\linewidth]{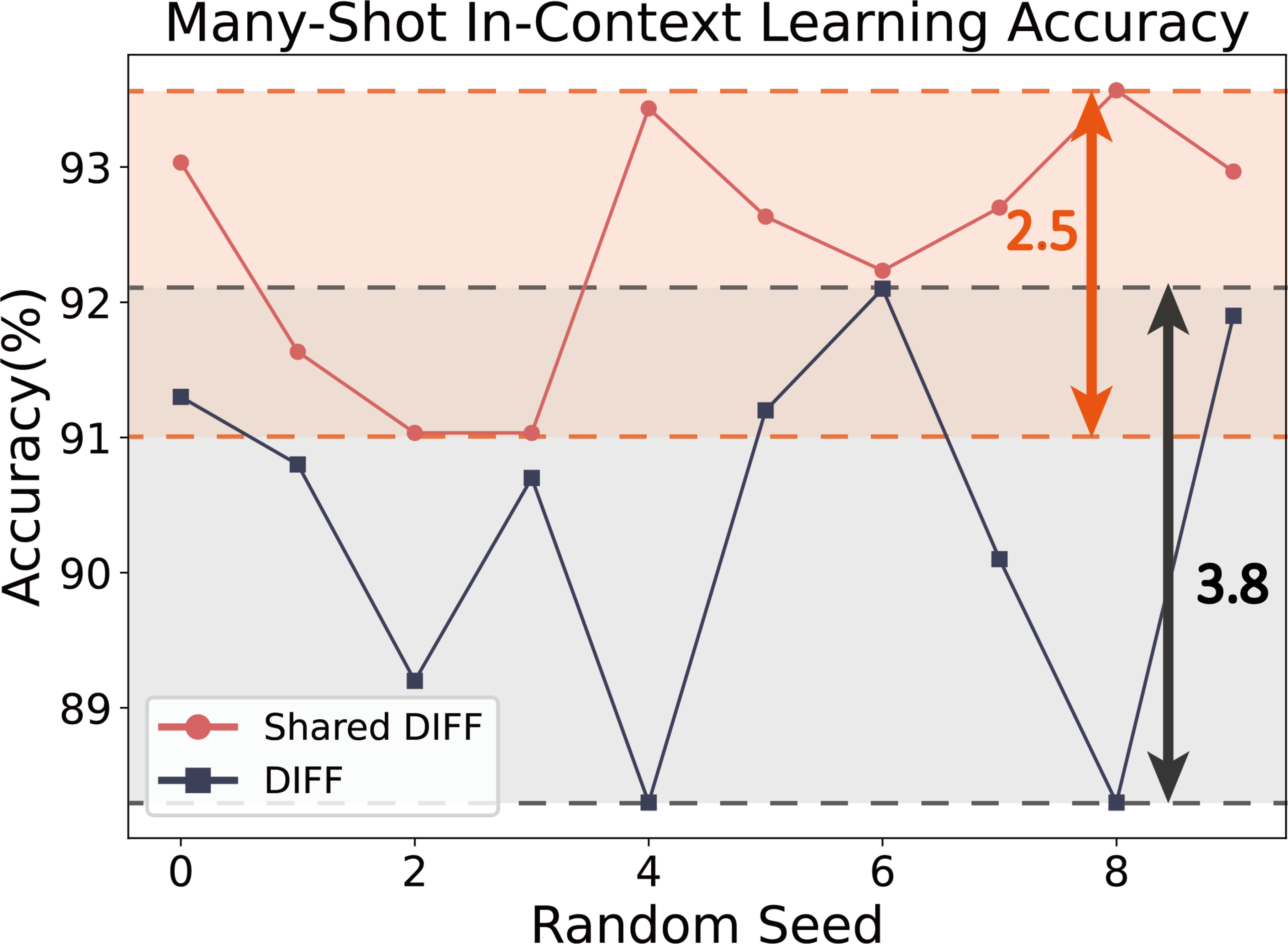}  
        \caption*{(a) Examples are randomly arranged.}
    \end{minipage}
    \hfill
    \begin{minipage}[b]{0.48\linewidth}
        \centering
        \includegraphics[width=\linewidth]{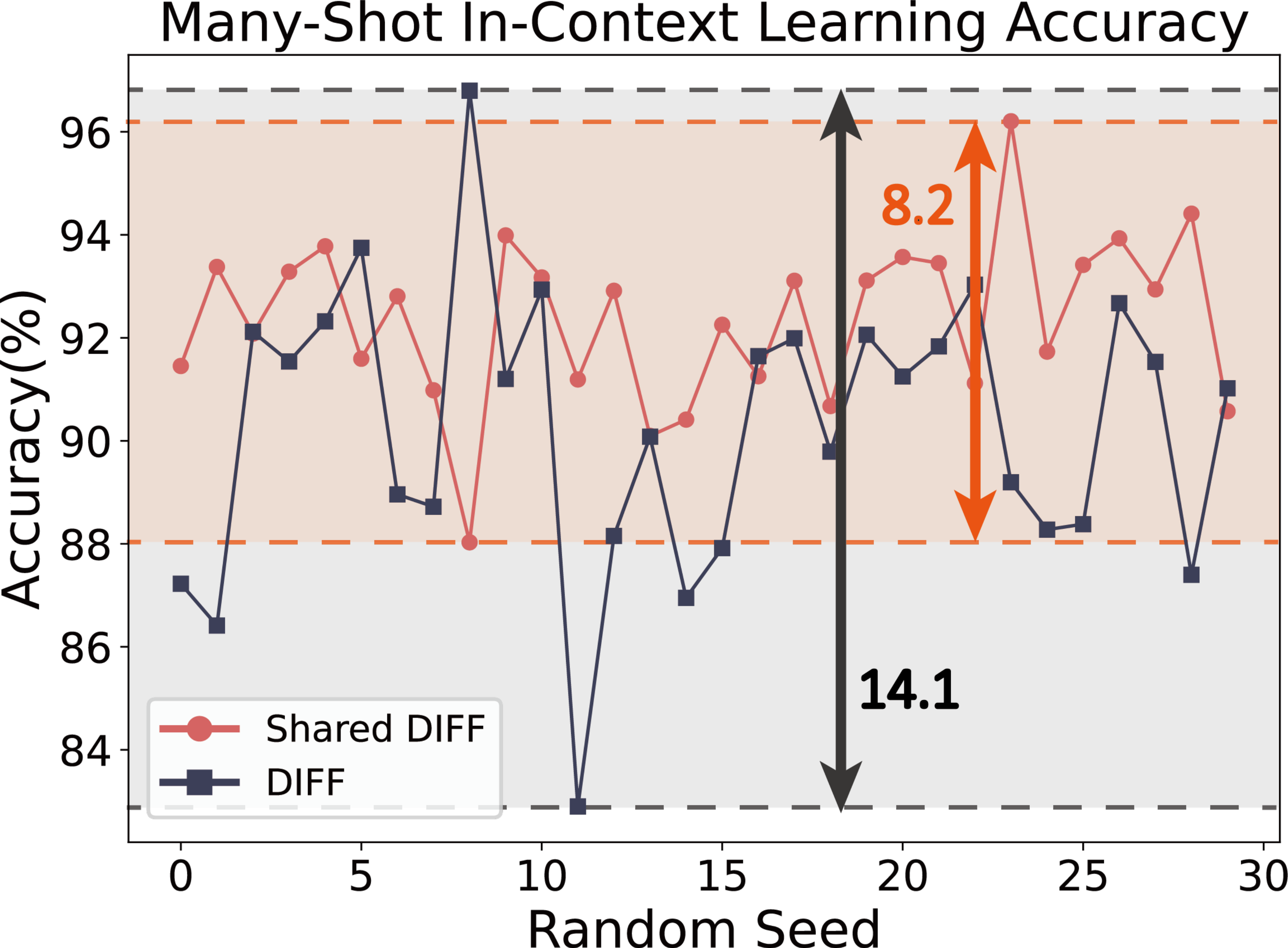}
        \caption*{(b) Examples are arranged alternately by class.}
    \end{minipage}
    \caption{Many-shot in-context learning accuracy on four datasets. The accuracy for both DIFF Transformer and DINT (Ours) models is presented, showing performance improvements across different numbers of demonstration samples.}
    \label{fig:Robustness}
\end{figure}

\begin{table*}[h]
\centering
\begin{tabular}{lccccccc} 
\toprule
\textbf{Model} & \textbf{\#Heads} & \textbf{d} & \textbf{GN} & \textbf{Valid. Set↓} & \textbf{AR-Hit↓} & \textbf{Others↓} \\
\midrule
DIFF  & 8  & 256 & \ding{51} & 3.062 & 0.880 & 3.247 \\
\textminus GroupNorm  & 8  & 128 & \ding{55} & 3.122 & 0.911 & 3.309 \\
with $\lambda_{\text{init}} = 0.8$  & 8  & 128 & \ding{51} & 3.065 & 0.883 & 3.250 \\
with $\lambda_{\text{init}} = 0.5$  & 8  & 128 & \ding{51} & 3.066 & 0.882 & 3.251 \\
\midrule
\textbf{Shared DIFF} & 8  & 128 & \ding{51} & \textbf{3.057} & \textbf{0.876} & \textbf{3.245} \\
\textminus GroupNorm & 8  & 128 & \ding{55} & 3.110 & 0.903 & 3.297 \\
with $\lambda_{\text{init}} = 0.8$  & 8  & 128 & \ding{51} & 3.060 & 0.881 & 3.246 \\
with $\lambda_{\text{init}} = 0.5$ & 8  & 128 & \ding{51} & 3.059 & 0.881 & 3.247 \\
\bottomrule
\end{tabular}
\caption{Evaluation of robustness in in-context learning on the TREC dataset. }
\label{tab:ablation_dint_diff}
\end{table*}

\textbf{Many-Shot In-Context Learning} As shown in Figure \ref{fig:multi_needle_results}, we compare the accuracy of Transformer, DIFF Transformer and Shared DIFF Transformer in many-shot classification tasks with 3B models and 64K input length. Following the evaluation protocol from \cite{bertsch2024in}, we gradually increase the number of demonstration samples to 64K tokens. Evaluating on datasets like TREC \cite{hovy2001toward}, TREC-Fine \cite{hovy2001toward}, Banking-77 \cite{casanueva2020efficient}, and Clinic-150 \cite{larson2019evaluation}, Shared DIFF Transformer outperforms DIFF Transformer with significant accuracy improvements: \textbf{2.0\% on TREC, 3.3\% on TREC-Fine, 3.1\% on Banking-77, and 1.5\% on Clinic-150.}

\textbf{Robustness of In-Context Learning} Figure \ref{fig:Robustness} compares the robustness of DIFF Transformer and Shared DIFF Transformer in in-context learning, analyzing performance fluctuations with different order permutations of demonstration examples. Shared DIFF Transformer consistently shows smaller fluctuations, indicating better robustness. In Figure \ref{fig:Robustness}.a, it reduces fluctuation by 34\%, and in Figure \ref{fig:Robustness}.b, by 42\%.

\subsection{Ablation Studies}

We conducted ablation studies with the same training setup as in Section 3.2 for the 1.4B model. Table~\ref{tab:ablation_dint_diff} reports the fine-grained loss on the validation set, divided into "AR-Hit" (recall of n-grams) and "Others" (frequent or unrecalled tokens).

As shown in Table~\ref{tab:ablation_dint_diff}, ablation studies on various design choices in DIFF and Shared DIFF Transformers reveal that GroupNorm significantly impacts performance, ensuring numerical stability by making the attention matrices sum to 1. We also tested different $\lambda$ initialization strategies, finding that the default method outperforms $\lambda_{\text{init}} = 0.8$ and 0.5, with minimal performance differences, demonstrating the robustness of the model to initialization variations.

\section{Conclusions}
\label{sec:typestyle}

In this study, inspired by differential amplifiers, we propose Shared DIFF Transformer, which effectively reduces parameter complexity by introducing shared base matrices and low-rank updates. Through a series of extensive experiments, we validate the advantages of Shared DIFF Transformer across multiple natural language processing tasks, including text classification, question answering, and sequence generation. The experimental results demonstrate that Shared DIFF Transformer not only improves accuracy and recall rates but also exhibits stronger robustness and scalability. These advantages position Shared DIFF Transformer as a promising approach for future applications in natural language processing.



\bibliographystyle{IEEEbib}
\bibliography{strings,refs}

\end{document}